\documentclass[letterpaper, 10 pt, conference]{ieeeconf}  %
\IEEEoverridecommandlockouts 
\usepackage{tabularx,booktabs}
\newcolumntype{C}{>{\centering\arraybackslash}X} 
\usepackage{multirow}
\usepackage{algorithm}
\usepackage{cite}
\usepackage{caption}
\usepackage{amsmath,amssymb,amsfonts}
\usepackage{graphicx}
\graphicspath{{Figure/}}
\usepackage{textcomp}
\usepackage{wrapfig}
\usepackage{amsmath,adjustbox,mathtools}
\usepackage{gensymb}
\usepackage{comment}
\usepackage{booktabs,caption}
\usepackage{multirow}
\usepackage{graphicx}
\usepackage{dblfloatfix}
\usepackage{textcomp}
\usepackage{algcompatible}
\usepackage[flushleft]{threeparttable}
\usepackage[dvipsnames]{xcolor}
\usepackage{booktabs}
\usepackage{xcolor,colortbl}
\usepackage{xcolor}
\usepackage{hyperref}
\usepackage[ruled,vlined,algo2e]{algorithm2e}
\usepackage{array,multirow}
\usepackage{lipsum}
\usepackage{balance}
\newcolumntype{P}[1]{>{\centering\arraybackslash}p{#1}}
\newcolumntype{M}[1]{>{\centering\arraybackslash}m{#1}}
\hypersetup{nolinks=true}
\makeatletter

\title{\LARGE \bf
3DS-SLAM: A 3D Object Detection based Semantic SLAM towards Dynamic Indoor Environments 
}

\author{
\authorblockN{
Ghanta Sai Krishna, Kundrapu Supriya, Sabur Baidya 
}
\thanks{
Ghanta Sai Krishna and Kundrapu Supriya are with the 
IIIT Naya Raipur, India (email: ghanta20102@iiitnr.edu.in, kundrapu20100@iiitnr.edu.in) 
}
\thanks{
Sabur Baidya is with the Department of Computer Science and Engineering, University of Louisville, USA; (e-mail: sabur.baidya@louisville.edu) 
}
\thanks{This work is partially supported by the US National Science Foundation (EPSCoR 1849213).
}}
\makeatletter
\pubid{\vspace{4mm}
\begin{minipage}{\textwidth}
        \centering\footnotesize{    Copyright~\copyright~2023 IEEE. Personal use of this material is permitted. Permission from IEEE must be obtained for all other uses, in any current or future media including reprinting/republishing this material for advertising or promotional purposes, creating new collective works, for resale or redistribution to servers or lists, or reuse of any copyrighted component of this work in other works.
}
\end{minipage}} 
  
\begin{document}
\maketitle
\vspace*{-10mm}
\begin{abstract}
The existence of variable factors within the environment can cause a decline in camera localization accuracy, as it violates the fundamental assumption of a static environment in Simultaneous Localization and Mapping (SLAM) algorithms.
Recent semantic SLAM systems towards dynamic environments either rely solely on 2D semantic information, or solely on geometric information, or combine their results in a loosely integrated manner. In this research paper, we introduce 3DS-SLAM, 3D Semantic SLAM, tailored for dynamic scenes with visual 3D object detection. The 3DS-SLAM is a tightly-coupled algorithm resolving both semantic and geometric constraints sequentially. We designed a 3D part-aware hybrid transformer for point cloud-based object detection to identify dynamic objects. Subsequently, we propose a dynamic feature filter based on HDBSCAN clustering to extract objects with significant absolute depth differences. When compared against ORB-SLAM2, 3DS-SLAM exhibits an average improvement of 98.01\% across the dynamic sequences of the TUM RGB-D dataset. Furthermore, it surpasses the performance of the other four leading SLAM systems designed for dynamic environments. The code and pretrained models are available at \href{https://github.com/sai-krishna-ghanta/3DS-SLAM}{\color{blue}{https://github.com/sai-krishna-ghanta/3DS-SLAM}}
\end{abstract}

\section{INTRODUCTION}
Simultaneous Localization and Mapping (SLAM) creates a map of its unknown surroundings while determining its own location using data from the sensors installed on the system. 
Although different sensors can contribute forming maps in SLAM systems, visual SLAM~\cite{chong2015sensor} is becoming increasingly popular for being able to produce fine-grained mapping information that is useful for many applications, e.g., robotics, transportation, search and rescue, constructions and many others.
Visual SLAM primarily relies on cameras of various types, encompassing monocular, stereo, and RGB-D cameras, due to their ability to comprehend scene compared to other sensors, e.g., lasers~\cite{fuentes2015visual}. Visual SLAM has undergone over three decades of continuous development, gradually maturing and proving its efficacy in static scenarios. Despite their strengths in controlled settings, traditional visual SLAM systems like ORB-SLAM2 \cite{mur2017orb}, LSD-SLAM \cite{engel2014lsd}, and RGBD-SLAM-V2 \cite{endres20133} can exhibit fragility when faced with challenging environments, such as dynamic or rough conditions.

In visual SLAM, object recognition is an inherent component for understanding the scene in the surroundings. In recent times, 
3D object detection has garnered significant interest as it aims for simultaneous localization and object recognition within a 3D point space. Being an essential foundation for comprehending semantic knowledge in indoor 3D space, 3D object detection can hold significant research gravity in Visual/ Semantic SLAM 
\cite{shu2021slam}. Thus, seamless integration of 3D object detection algorithms with Visual SLAM is a pivotal research direction with far-reaching implications. 

\begin{figure}[t]
    \centering
    \centerline{\includegraphics[width=\linewidth]{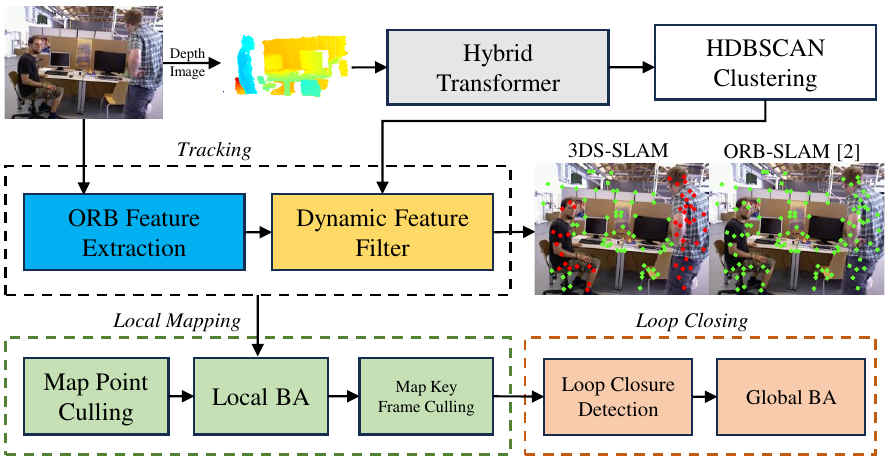}}
    \captionsetup{font=small}
    \caption {\footnotesize{\textbf{3DS-SLAM - Overview:} A 3D visual SLAM system for indoor dynamic environments. Existing ORB-SLAM2 fails due to dynamic features on moving people, rendering the estimated trajectory unusable. 3DS-SLAM employs HTx architecture for 3D object detection and leverages HDBSCAN to extract dynamic features (red points) and improves overall stability.}}
    \label{fig:introduction}
    \vspace{-6mm}
\end{figure} 

However, in presence of dynamic objects, traditional Visual SLAM systems face challenges in accuracy and robustness, mainly due to issues with the data association caused by dynamic points in images. Researchers have sought solutions to mitigate these challenges, leveraging deep learning technology to address Visual SLAM in dynamic environments. These approaches utilize techniques, e.g., 2D object detection \cite{zhong2018detect}, semantic segmentation \cite{fan2022blitz}, resolving geometric constraints (e.g., RANSAC \cite{ebrahimi2021automatic}, DBSCAN\cite{chen2022approach}), and epipolar/projection constraints \cite{hu2022cfp}. Traditional methods combine geometric and semantic information, employing two voting strategies --  (i) if both the semantic and geometric modules identify a feature as dynamic, it is classified as dynamic \cite{8593691}; and (ii) if at least one module identifies it as dynamic, it is considered dynamic \cite{9385844}. The visual SLAM systems with these voting strategies are often considered as loosely coupled SLAM systems \cite{cui2019sof}. 

Among deep learning based methods, semantic segmentation offers precise object masks at the pixel level, while maintaining real-time performance. However, it is constrained by high computational costs and potential in- \\ \\ \\ accuracies in capturing moving objects. In contrast, 2D object detection, while faster with bounding boxes, may struggle with background noise and complex cases. For foreground-background separation, the RANSAC algorithm works well in static or mildly dynamic environments but struggles when highly dynamic objects become predominant in the camera's field of view.

To address these challenges, we present 3DS-SLAM, a high-performance and the first 3D visual SLAM system optimized for dynamic indoor settings. Built upon the foundation of ORB-SLAM2\cite{mur2017orb}, 3DS-SLAM proposed Hybrid Transformer architecture (HTx) for semantic information (3D object dettection) and uses HDBSCAN (Hierarchical Density-Based Spatial Clustering) for resolving geometric constraints with HTx results as shown in Fig. \ref{fig:introduction}. Using 3D object detection over 2D object detection for SLAM provides improved spatial understanding, better occlusion handling, accurate scale estimation, and enhanced motion tracking capabilities. By conducting comprehensive experiments on publicly available datasets, we demonstrate that our approach outperforms the current state-of-the-art methods (SOTA) dynamic visual SLAM methods, demonstrating superior localization accuracy across high dynamic scenarios.

\textbf{The summary of our contributions are as follows:}

\begin{itemize}
    \item A lightweight 3D HTx object detection architecture integrating our visual SLAM system, enabling 3D semantic-spatial information for dynamic environments.
    \item A novel end-to-end pipeline integrating HTx and HDBSCAN, which effectively addresses both semantic and geometric constraints, optimizing overall performance.
    \item Experimental validation shows that 3DS-SLAM enhances pose accuracy and stability in dynamic scenes, outperforming existing methods.
\end{itemize}

\section{BACKGROUND/ RELATED WORK}
\subsection{Visual SLAM with Semantic \& Geometric Constraints}
In context of our work, we will discuss ORB-SLAM2 based existing state-of-art frameworks addressing both semantic and geometric constraints. CFP-SLAM \cite{hu2022cfp} incorporates 2D object detection and combines semantic and coarse-to-fine static probability-based epipolar geometric information to estimate camera poses. However, the CFP-SLAM fails in RGB sequences with huge change in camera rotation, due to the dis-functionality in epipolar constraint method. Another approach, namely SaD-SLAM \cite{yuan2020sad} extends ORB-SLAM2 with semantic-depth-based movable object tracking and enhanced camera pose estimation through Mask RCNN-based \cite{he2017mask} feature point fusion in dynamic environments. SOF-SLAM \cite{cui2019sof}, a semantic SLAM system tailored for dynamic environments. SOF-SLAM uses semantic optical flow and SegNet's pixel-wise segmentation \cite{badrinarayanan2015segnet} to ensure precise estimation of camera poses in dynamic environments utilizing static features. In DS-SLAM \cite{yu2018ds}, semantic information is acquired through SegNet \cite{badrinarayanan2015segnet}, incorporating sparse optical flow and motion consistency analysis to differentiate dynamic and static characteristics of individuals. Dyna-SLAM \cite{9385844}, on the other hand, integrates mask R-CNN and multi-view geometry techniques to handle dynamic elements. In YOLO-SLAM \cite{wu2022yolo}, Darknet19-YOLOv3 \cite{redmon2018yolov3} and a novel depth-based geometric constraint method are combined to efficiently reduce the influence of dynamic features.

\subsection{3D Object Detection and Transformers}
In 3D object detection, three feature categories are distinguished in architectural utilization \cite{li2022unifying}: (i) point features, (ii) voxel features, and (iii) point-voxel features. Recent advancements in transformer architectures have prompted efforts to work on both point and voxel object detection. 3-DETR \cite{misra2021end}, a typical transformer architecture incorporating non-parametric queries and fourier positional embeddings achieved a 9.5\% performance improvement over highly optimized methods like VoteNet \cite{qi2019deep} on ScanNetV2 \cite{dai2017scannet} and SUNRGB-D \cite{song2015sun} datasets. PVT-SSD \cite{yang2023pvt} proposed a hybrid approach with both point and voxel feature representations. This architecture leverages voxel-based sparse convolutions to perform feature extraction, combined with the Point-Voxel Transformer (PVT) for 3D object detection. In HVNet \cite{ye2020hvnet} , the authors proprosed to use multi-scale voxel feature encoder to extract the features, and then a dynamic feature projection and convolutional backbone network for prediction. Early research works primarily combined point and voxel features when inputting data into the architecture. However, to bolster the resilience of 3D object detection in existing research works, certain factors such as point cloud and algorithm complexities, as well as real-time constraints, have been largely left unaddressed. HTx introduces a hybrid framework that leverages class-aware 3D object detection utilizing features from raw points and part-aware 3D object detection using voxel features. Hence, this work enhances the perceptual capabilities of robots by developing a HTx based visual SLAM capable of comprehending a 3D scene.  

\section{3DS-SLAM: THE APPROACH}
\subsection{System Architecture}

\begin{figure*}[ht]
    \centering
    \centerline{\includegraphics[width=0.90\linewidth]{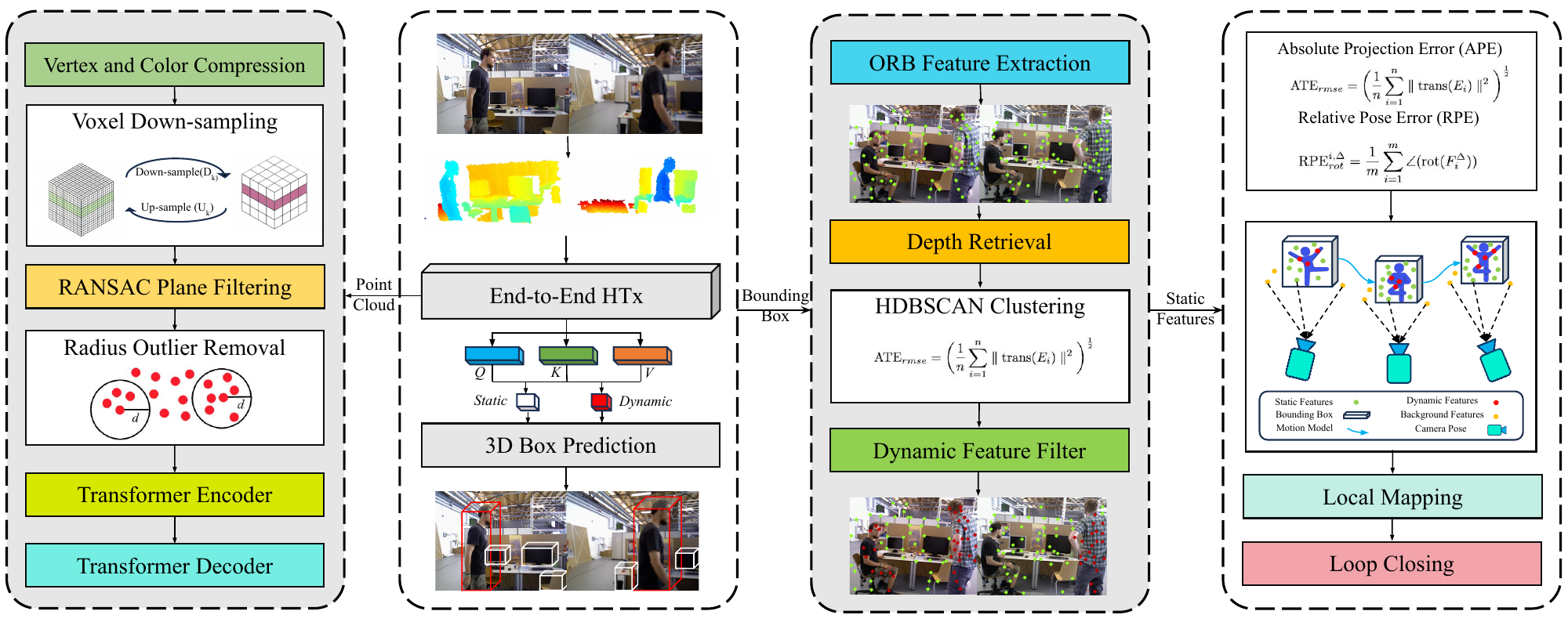}}
    \vspace{-2mm}
    \caption{{\textbf{3DS-SLAM - Framework:} it is subdivied into mainly three components: 1.) 3D object detection thread. 2.) Dynamic feature removal thread. 3.) tracking, local mapping, local closing threads adapted from ORB-SLAM2.}}
    \label{fig:concept}
    \vspace{-7mm}
\end{figure*} 

The proposed 3DS-SLAM extends the capabilities of ORB-SLAM2, originally designed for static environments, by incorporating two additional threads -- {\it 3D object detection} and {\it dynamic features filter} as shown in Fig. \ref{fig:concept}. These threads effectively filter dynamic points, ensuring precise camera trajectory estimation. For semantic information, the {\it 3D object detection} thread employs a light-weight HTx arhitecture, while the {\it dynamic features filter} thread utilizes geometrical depth-based HDBSCAN clustering to distinguish dynamic points. The system utilizes HTx architecture to extract semantic information from point clouds extracted from RGB and depth images.

\subsection{Hybrid Transformer: Light-Weight 3D Object Detector}
In visual SLAM, the frames captured by the sensor often exhibit incomplete foreground objects, which can lead to compromised object detection. This necessitates the development of partial object localization methods that are aware of these incomplete object part representations. The proposed HTx architecture takes input as 3D point clouds to predict object positions, encompassing depth, orientation, and position of object. Our proposed HTx architecture is designed based on the building blocks from \cite{shi2020points} for part-aware object localization and \cite{misra2021end} for class-aware object localization. The HTx architecture differs from existing transformer architectures at the data level in terms of incorporating point-cloud preprocessing and utilizing both point and voxel features for part-aware object localization.

A point cloud consists of a disordered set of $N$ points, each seamlessly tied to its 3-dimensional XYZ coordinates. Due to their increased computational complexity when compared to images, this research diligently conducts extensive preprocessing to effectively compress the point clouds.
Moreover, the inherent permutational invariance of point clouds, combined with the inclusion of color information and point normals, also result in substantial computational overhead for 3D object detection. Taking inspiration from prior work \cite{qi2019deep}, our HTx architecture prioritizes real-time efficiency by forgoing the use of color and point normals information for object detection. Furthermore, significant data preprocessing techniques such as voxel down-sampling, plane filtering, radius based outiler removal have been performed.

\subsubsection{\textbf{Voxel Downsampling}}Let $\mathcal{P} = \{(x_i, y_i, z_i)\}_{i=1}^N$ be the point cloud with $N$ points in 3D space. The voxel downsampling process involves defining a voxel size $\Delta x \times \Delta y \times \Delta z$ and associating each point with its corresponding voxel $(i, j, k)$ using the floor function: $i = \left\lfloor \frac{x_i}{\Delta x} \right\rfloor$, $j = \left\lfloor \frac{y_i}{\Delta y} \right\rfloor$, and $k = \left\lfloor \frac{z_i}{\Delta z} \right\rfloor$. The downsampled point cloud is then represented by the centroids $C_{ijk}$ of each voxel, this is determined by computing the coordinate average for all points within that voxel, resulting point cloud $\mathcal{P'} = \{(x'_i, y'_i, z'_i)\}_{i=1}^{N^ {\!\!\!\!\!'}}$ . 

\subsubsection{\textbf{RANSAC-Ground Filtering}} The objective is to find a plane represented by the equation $ax + by + cz + d = 0$ that best fits a subset of points. The RANSAC iteratively randomly selects minimal sets of points to estimate the plane parameters, forming a consensus set of inliers within a distance threshold, and selects the plane with the largest consensus set as the final best-fitting ground plane.

\subsubsection{\textbf{Radius based Outlier Removal}} From the preprocessed point cloud $\mathcal{P''}$ and user-defined radius $R$, this iterative algorithm identifies points with fewer neighbors within this radius as outliers. For each point $(x''_i, y''_i, z''_i)$, it computes the number of points $n_i$ within the radius $R$ centered at  $(x''_i, y''_i, z''_i)$. If $n_i$ is below a threshold, the point is considered an outlier and  excluded from the point cloud.

\subsubsection{\textbf{Transformer Architecture}} 
The 3DS-SLAM employs a framework for 3D object detection, built upon the pioneering 3DETR architecture by Facebook AI Research \cite{misra2021end}. We have significantly modified this architecture to enhance its compatibility with visual SLAM systems, particularly by reinforcing the part-aware object detection layer introduced in \cite{shi2020points}. This addresses a crucial gap in existing visual SLAM systems, which does not address object detection in critical robotic applications due to partially visible objects, camera rotation and other environmental factors. Due to the complexity of designing a loss function for both part-aware and class-aware object localization, we have developed two separate loss functions.

The prediction MLPs (Multi-Layer Perceptron)  generate a 3D bounding box \( \hat{b} \), which is further evaluated with actual box \( b \). Each predicted box \( \hat{b} = [ \hat{c}, \hat{d}, \hat{a}, \hat{s} ] \) includes (1) geometric elements \( \hat{c}, \hat{d} \in [0, 1]^3 \) that define the box's center and dimensions, \( \hat{a} = [ \hat{a}_c, \hat{a}_r ] \) representing the class and orientation residue, and (2) the semantic term \( \hat{s} = [0, 1]^{K+1} \) containing the probability distribution over \( K \) semantic object classes and the 'background' class. We employed $ \ell_{1} $ regression losses for center and box dimensions, along with Huber regression loss \cite{tong2023functional} for angular residuals, and cross-entropy losses for angular and semantic classifications as follows:
\vspace{-3mm}
\begin{equation}
\begin{aligned}    \mathcal{L}_{\mathrm{}}=\lambda_c\|\hat{\mathbf{c}}-\mathbf{c}\|_1+\lambda_d\|\hat{\mathbf{d}}-\mathbf{d}\|_1 +  \\ \lambda_{a r}\left\|\hat{\mathbf{a}}_r-\mathbf{a}_r\right\|_{\text {huber }}
-\lambda_{a c} \mathbf{a}_c^{\top} \log \hat{\mathbf{a}}_c-\lambda_s \mathbf{s}_c^{\boldsymbol{T}} \log \hat{\mathbf{s}}_c
\end{aligned}
\end{equation}

\begin{figure*}[!b]
    \centering
    \vspace{-4mm}
\centerline{\includegraphics[width=\linewidth]{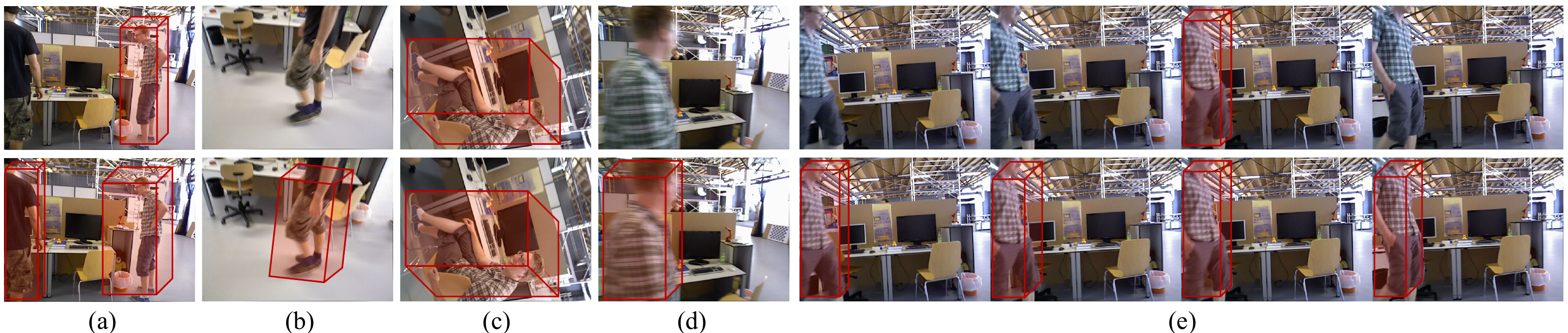}}
\vspace{-2mm}    \caption{The results of 3D object detection with and without part-aware object localization in the following cases: (a) The limited visibility of the object within the camera's field of view (b) The swift motion of the dynamic object. (c) The singular viewpoint from camera rotation. (d) The blurred image.  (e) Continuous frame partial object detection.}
    \label{fig:objectdetectionresults}
\end{figure*} 

\subsubsection{\textbf{Part-aware Object Localization}} 
We define the intra-object part location for each point by representing it as its relative position within the 3D bounding box of the ground-truth object to which it is assigned. We represent this target intra-object part 
location using three continuous values, denoted as ($x^{f}$, $y^{f}$, $z^{f}$), for each point ($x^{p}$, $y^{p}$, $z^{p}$)~as:

\vspace*{-10pt}

\begin{equation}
\begin{aligned}
[x^{t}  \quad   y^{t}] =\left[\begin{array}{ll}
x^{p}-x^{c} & y^{p}-y^{c}
\end{array}\right]\left[\begin{array}{cc}
\cos\theta & \sin\theta \\
-\sin\theta & \cos\theta
\end{array}\right]\\
x^{f} =\frac{x^{t}}{w}+x^{c}, \quad y^{f}= \frac{y^{t}}{l}+y^{c},  \quad z^{f} =\frac{z^{p}-z^{c}}{h}+z^{c}
\end{aligned}
\end{equation}

where ($x^{c}$, $y^{c}$, $z^{c}$) represents the center of ground-truth bounding box, denoting its position in 3D space. The box size and orientation is represented by ($h$, $w$, $l$, $\theta$), corresponding to its height, width, top-view angle. ($x^{t},  y^{t}$) are considered as temporary variables for each point iteration.  To estimate the intra-object part location for each point, represented as ($x^{f}$, $y^{f}$, $z^{f}$), we employ binary cross-entropy loss \cite{ho2019real} for each point , defined as follows:
\begin{equation}
\begin{aligned}
\mathcal{L}_{f}\left(u^{f}\right)= -u^{f} \log \left(\tilde{u}^{f}\right) 
& -\left(1-u^{f}\right) \log \left(1-\tilde{u}^{f}\right) \\ 
\end{aligned}
\end{equation}

where $\tilde{u}^{f}$ represents the predicted intra-object part location, $u^{f}$ denotes the corresponding actual intra-object part location and $u \in\{x, y, z\}$.

\subsection{HDBSCAN Clustering and Dynamic Feature Filter}
Object detection methods may not accurately provide
object masks, especially when dealing with non-rigid bodies
that fill a substantial portion of the camera's field of view. This often results in numerous background point clouds
being included within the object’s bounding box. To address
this issue, we focus on human subjects as an example
of non-rigid foreground bodies. Humans typically exhibit
depth continuity and significant depth disparity from the
background. Therefore, when a person’s bounding box dominates the camera’s view, we optimize the native HDBSCAN density clustering algorithm to differentiate between points in the foreground and those in the background within the bounding box. By combining groups of points with shallow depths, we identify
the foreground (dynamic keypoints). This adaptive approach
enhances the robustness of the HDBSCAN algorithm and
effectively handles cases where people are partially occluded by other objects. Furthermore, the HDBSCAN is more robust to parameter selection and handles varying density multidimensional data effectively compared to DBSCAN \cite{chen2022approach}.

The HDBSCAN algorithm is utilized to process 3D spatial keypoints $K$ and depth map data $D$, creating a space of points denoted as $I_{(k_{x}, K_{y}, d)}$. Within HDBSCAN, point density is defined by the core distance ${k}(i)$, which represents the Euclidean distance to the k-th nearest neighbor of a point. To distinguish low-density points (high core distance), a distance metric called mutual reachability distance ($d_{m r}$) as shown in eq. \ref{eq:hdbscan} is utilized. 
\vspace{-5mm}
\begin{equation}
d_{m r}(i, j)= \begin{cases}\max \{\kappa(i), \kappa(j), d(i, j)\}, & i \neq j \\ 0, & i=j\end{cases}
\label{eq:hdbscan}
\end{equation}

The clusters are extracted based on cluster stability and persistence defined using $\lambda$ values. The proposed modifications of HDBSCAN is efficient in extracting 2 clusters for foreground and background, representing dynamic and static features respectively with keypoints and depth map.

\begin{figure*}[t]
    \centering
    \centerline{\includegraphics[width=\linewidth]{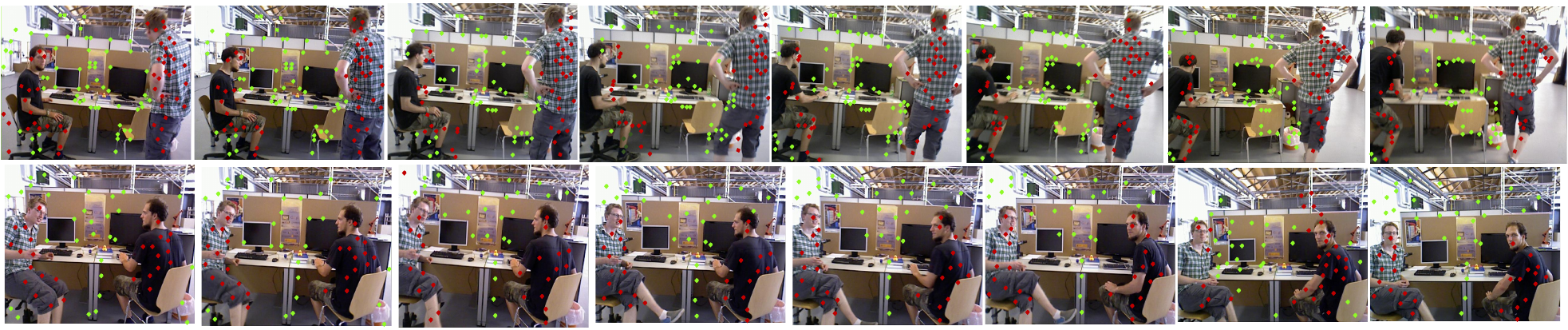}}
    \caption{The overall results of 3DS-SLAM in two sets of consecutive, leaving 4 frames in the between.}
    \label{fig:hdscanoverall}
    \vspace{-4mm}
\end{figure*}

\section{EXPERIMENTAL ANALYSIS AND RESULTS}
We conducted experiments to evaluate 3DS-SLAM in 8 dynamic sequences extracted from the TUM RGB-D dataset \cite{sturm12iros}, comprising 4 sitting (fr3/s) and 4 walking (fr3/w) sequences, with camera motions including static, xyz, halfsphere, and roll-pitch-yaw (rpy). We compare 3DS-SLAM to a naive semantic ORB-SLAM2 system, demonstrating our approach's superior effectiveness. Subsequently, we evaluate our method against SOTA SLAM systems in dynamic environments, providing valuable insights. We also showcase our system's capabilities in a controlled laboratory environment. These experiments utilize a computer with Ubuntu 20.04, an i9 CPU, 16GB RAM, and RTX 3070 Ti GPU. Furthermore, 3DS-SLAM considers Absolute Trajectory Error (ATE) and Relative Pose Error (RPE) as metrics to evaluate the performance. The ATE quantifies the global trajectory accuracy whereas RPE measures local consistency over a fixed time interval. Initially, we analyze the performance of the 3D object detection and geometric depth filter then compare 3DS-SLAM stability and robustness with existing works. A relative comparison is carried out using Root-Mean-Square-Error (RMSE) and Standard Deviation (S.D.) of both Absolute Trajectory Error (ATE) and Relative Pose Error (RPE) \cite{prokhorov2019measuring} to evaluate the performance of 3DS-SLAM.

\begin{figure*}[t]
    \centering
\centerline{\includegraphics[width=\linewidth]{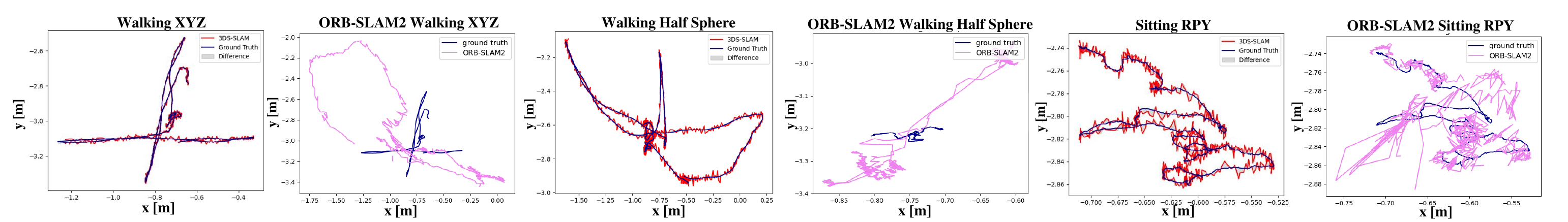}}
    \vspace{-2mm}\caption{The trajectory comparison between 3DS-SLAM and ORB-SLAM2 high and low dynamic sequences }
    \label{fig:Plot_Results}
\end{figure*}

\begin{table*}[]
    \centering 
    \captionsetup{font=footnotesize}
    \caption{COMPARISON OF ABSOLUTE TRAJECTORY ERROR (ATE) WITH EXISTING ARCHITECTURES}
    \vspace{-10pt}
    \begin{center}
    \resizebox{\textwidth}{!}{
    {\renewcommand{\arraystretch}{1.3} 
    \begin{tabular}{M{1.5cm}|cc|cc|cc|cc|cc|cc|cc}
\hline
\multirow{2}{*}{\hfil Sequences}   & \multicolumn{2}{c}{\text { ORB-SLAM2 }} & \multicolumn{2}{c}{\text { YOLO-SLAM }} & \multicolumn{2}{c}{\text { DS-SLAM }} & \multicolumn{2}{c}{\text { DYNA-SLAM }} & \multicolumn{2}{c}{\text { RDS-SLAM }} & \multicolumn{2}{c}{\text { CFP-SLAM }} & \multicolumn{2}{c}{\text { 3DS-SLAM }} 
 \\\cline{2-15}& 

RMSE  & SD &RMSE  & SD & RMSE  & SD & RMSE  & SD & RMSE  & SD &  RMSE  & SD  &  RMSE  & SD \\

\hline \vspace{1pt}
fr3/w/xyz & 0.7214 & 0.2560 & 0.0146 & 0.0070 & 0.0247 & 0.0161 & 0.0164 & 0.0086 & 0.0240 & 0.0139 & 0.0141 & 0.0072 & 
\textbf{0.0135} & \textbf{0.0070} \\
 
fr3/w/half & 0.4667 & 0.2601 & 0.0283 & 0.0138 & 0.0303 & 0.0159 & 0.0296 & 0.0157 & 0.0306 & 0.0171 & 0.0237 & 0.0114 &  \textbf{0.0197} & \textbf{0.0105}  \\
 
fr3/w/static & 0.3872 & 0.1632 & 0.0073 & 0.0035 & 0.0081 & 0.0036 & 0.0068 & 0.0032 & 0.0720 & 0.0343 & 0.0066 & 0.0030 & 
\textbf{0.0063} & \textbf{0.0025} \\
 
fr3/w/rpy & 0.7842 & 0.4005 & 0.2164 & 0.1001 & 0.4442 & 0.2350 & \textbf{0.0354} & \textbf{0.0190} & 0.0587 & 0.0380 & 0.0368 & 0.0230 &  0.0370 & 0.0245 \\
 
fr3/s/xyz & 0.0092 & 0.0047 & --- & --- & --- & --- & 0.0127 & 0.0060 & --- & --- & 0.0090 & 0.0042 & \textbf{0.0085} & \textbf{0.0040} \\
 
fr3/s/half & 0.0192 & 0.0110 & --- & --- & --- & --- & 0.0186 & 0.0086 & --- & --- & 0.0147 & 0.0069 & \textbf{0.0135} & \textbf{0.0065} \\
 
fr3/s/static & 0.0087 & 0.1636 & 0.0066 & 0.0033 & 0.0065 & 0.0033 & --- & --- & 0.0084 & 0.0043 & 0.0053 & 0.0027 & \textbf{0.0047} & \textbf{0.0021} \\
 
fr3/s/rpy & 0.0195 & 0.0124 & --- & --- & --- & --- & --- & --- & --- & --- & 0.0253 & \textbf{0.0154} & \textbf{0.0252} & 0.0158 \\
 
\hline
    \end{tabular}}}
    \end{center}
    \vspace*{-7pt}
    \begin{tablenotes}
      \small
      \item \footnotesize{--- represents corresponding data is not mentioned in the respective literature. The text in bold indicates the scheme that outperformed all others.}
    \end{tablenotes}
    \label{tab:my_label}
    \vspace*{-12pt}
\end{table*}

\subsection{Performance of 3D Object Detection and HDBSCAN}
The 3DS-SLAM is developed addressing critical robotic applications with  point-cloud-based hybrid 3D object detection approach. The HTx architecture is trained on the SUNRGB-D dataset \cite{song2015sun}, which contains $700$ labeled objects in indoor environments, with consideration for future implications in standard industrial robotic applications such as reach, grasp and pick-and-place. The number of object categories in SUNRGB-D is nearly nine times greater than the categories in the coco dataset \cite{lin2014microsoft} utilized by YOLO \cite{redmon2018yolov3}. The HTx architecture achieves an $mAP_{25}$ value \cite{padilla2020survey} of $57.85$, which is comparable to the top-performing 3D object detection models. Additionally, its tight integration with the dynamic feature filter enhances overall visual SLAM performance. The Figures \ref{fig:objectdetectionresults}(a)-(e) visually depict the results of class-aware HTx in various scenarios, comparing its performance with part-aware object localization (2nd row). The part-aware object localization significantly enhanced performance of 3DS-SLAM, particularly evident in the TUM RGB-D sequences. The 3D object detection results remain consistently accurate throughout all frames, resulting in a significantly smoother 3DS-SLAM experience compared to existing 2D object detection architectures that lacks missed detection compensation across multiple frames. The Fig. \ref{fig:hdscanoverall} represents the comprehensive results of 3DS-SLAM showing temporal continuous detection. The 3D bounding boxes and dynamic points are represented in red whereas static points are represented in green. 

\begin{table*}[]
    \centering 
    \captionsetup{font=footnotesize}
    \caption{COMPARISON OF TRANSLATIONAL RELATIVE POSE ERROR (RPE) WITH EXISTING ARCHITECTURES }
    \vspace{-10pt}
    \begin{center}
    \resizebox{\textwidth}{!}{
    {\renewcommand{\arraystretch}{1.3} 
    \begin{tabular}{M{1.5cm}|cc|cc|cc|cc|cc|cc|cc}
\hline
\multirow{2}{*}{\hfil Sequences}   & \multicolumn{2}{c}{\text { ORB-SLAM2 }} & \multicolumn{2}{c}{\text { YOLO-SLAM }} & \multicolumn{2}{c}{\text { DS-SLAM }} & \multicolumn{2}{c}{\text { DYNA-SLAM }} & \multicolumn{2}{c}{\text {RDS-SLAM }} & \multicolumn{2}{c}{\text { CFP-SLAM }} & \multicolumn{2}{c}{\text { 3DS-SLAM }} 
 \\\cline{2-15}& 

RMSE  & SD &RMSE  & SD & RMSE  & SD & RMSE  & SD & RMSE  & SD &  RMSE  & SD  &  RMSE  & SD \\

\hline \vspace{1pt}
fr3/w/xyz & 0.3944 & 0.2964 & 0.0194 & 0.0097 & 0.0333 & 0.0229 & 0.0217 & 0.0119 & 0.0299 & 0.4943 & 0.0190 & 0.0097 & \textbf{0.0183} & \textbf{0.0085} \\
 
fr3/w/half & 0.3480 & 0.2859 & 0.0268 & 0.0124 & 0.0297 & 0.0152 & 0.0284 & 0.0149 & 0.0332 & 0.0208 & 0.0259 & 0.0128 & \textbf{0.0247} & \textbf{0.0119} \\
 
fr3/w/static & 0.2349 & 0.2151 & 0.0094 & 0.0044 & 0.0102 & 0.0048 & 0.0089 & 0.0044 & 0.0529 & 0.0444 & 0.0089 & 0.0040 & \textbf{0.0078} & \textbf{0.0039} \\
 
fr3/w/rpy & 0.4582 & 0.3447 & 0.0933 & 0.0736 & 0.1168 & 0.0473 & \textbf{0.0448} & \textbf{0.0262} & 0.0700 & 0.0488 & 0.0500 & 0.0306 & 0.0511 & 0.0341 \\
 
fr3/s/xyz & 0.0117 & 0.0060 & --- & --- & --- & --- & 0.0142 & 0.0073 & --- & --- & \textbf{0.0114} & 0.0055 & 0.0120 & \textbf{0.0056} \\
 
fr3/s/half & 0.0231 & 0.0163 & --- & --- & --- & --- & 0.0239 & 0.0120 & --- & --- & 0.0162 & 0.0079 & \textbf{0.0143} & \textbf{0.0069} \\
 
fr3/s/static & 0.0090 & 0.0043 & 0.0089 & 0.0044 & 0.0078 & 0.0038 & --- & --- & 0.0097 & 0.0052 & 0.0072 & 0.0035 & \textbf{0.0068} & \textbf{0.0031} \\
 
fr3/s/rpy & 0.0245 & 0.0144 & --- & --- & --- & --- & --- & --- & --- & --- & \textbf{0.0316} & \textbf{0.0186} & 0.0320 & 0.0190 \\
 
\hline
    \end{tabular}}}
    \end{center}
    \vspace*{-12pt}
    \label{tab:my_label}
\end{table*}

\begin{table*}[]
    \centering 
    \captionsetup{font=footnotesize}
    \caption{COMPARISON OF ROTATIONAL RELATIVE POSE ERROR (RPE) WITH EXISTING ARCHITECTURES}
    \vspace{-10pt}
    \begin{center}
    \resizebox{\textwidth}{!}{
    {\renewcommand{\arraystretch}{1.3} 
    \begin{tabular}{M{1.5cm}|cc|cc|cc|cc|cc|cc|cc}
\hline
\multirow{2}{*}{\hfil Sequences}   & \multicolumn{2}{c}{\text { ORB-SLAM2 }} & \multicolumn{2}{c}{\text { YOLO-SLAM }} & \multicolumn{2}{c}{\text { DS-SLAM }} & \multicolumn{2}{c}{\text { DYNA-SLAM }} & \multicolumn{2}{c}{\text { RDS-SLAM }} & \multicolumn{2}{c}{\text { CFP-SLAM }} & \multicolumn{2}{c}{\text { 3DS-SLAM }} 
 \\\cline{2-15}& 

RMSE  & SD &RMSE  & SD & RMSE  & SD & RMSE  & SD & RMSE  & SD &  RMSE  & SD  &  RMSE  & SD \\

\hline \vspace{1pt}
fr3/w/xyz & 7.7846 & 5.8335 & 0.5984 & \textbf{0.3655} & 0.8266 & 0.5826 & 0.6284 & 0.3848 & 0.7739 & 0.4943 & 0.6023 & 0.3719 & \textbf{0.5819} & 0.3695 \\
 
fr3/w/half & 7.2138 & 5.8299 & 0.7534 & 0.3564 & 0.8142 & 0.4101 & 0.7842 & 0.4012 & 0.8194 & 0.4858 & 0.7575 & 0.3743 & \textbf{0.7511} &\textbf{0.3501}  \\
 
fr3/w/static & 4.1856 & 3.8077 & 0.2623 & 0.1104 & 0.2690 & 0.1182 & 0.2612 & 0.1259 & 1.4966 & 1.2839 & 0.2527 & 0.1051 & \textbf{0.2491} & \textbf{1011} \\
 
fr3/w/rpy & 8.8923 & 6.6658 & 1.8238 & 1.4611 & 3.0042 & 2.3065 & \textbf{0.9894} & \textbf{0.5701} & 1.4736 & 1.062 & 1.1084 & 0.6722 & 0.9901 &  0.5719 \\
 
fr3/s/xyz & 0.4890 & 0.2713 & --- & --- & --- & --- & 0.5042 & 0.2651 & --- & --- & 0.4875 & 0.2640 & \textbf{0.4866} & \textbf{0.2621} \\
 
fr3/s/half & 0.6015 & 0.2924 & --- & --- & --- & --- & 0.7045 & 0.3488 & --- & --- & 0.5917 & 0.2834 & \textbf{0.5899} & \textbf{0.2809} \\
 
fr3/s/static & 0.2850 & 0.1241 & 0.2709 & 0.1209 & 0.2735 & 0.1215 & --- & --- & 0.3217 & 0.1522 & 0.2654 & 0.1183 & \textbf{0.2609} & \textbf{0.1180} \\
 
fr3/s/rpy & 0.7772 & 0.3999 & --- & --- & --- & --- & --- & --- & --- & --- & 0.7410 & 0.3665 & \textbf{0.7399} & \textbf{0.3664} \\
 
\hline
    \end{tabular}}}
    \end{center}
    \vspace*{-12pt}
    \label{tab:my_label}
\end{table*}

\subsection{SLAM Peroformance Comparison with SOTA frameworks}
We conducted a comparative analysis between our approach and several state-of-the-art dynamic SLAM methods, including ORB-SLAM2 \cite{mur2017orb}, YOLO-SLAM \cite{wu2022yolo}, DS-SLAM \cite{8593691}, DYNA-SLAM \cite{8421015}, RDS-SLAM \cite{9318990}, and CFP-SLAM \cite{hu2022cfp} which also have demonstrated superior performance when compared to ORB-SLAM2. The quantitative evaluation results can be found in Tables I, II, and III, which present ATE, translational RPE, and rotational RPE across all eight TUM RGB-D sequences. In rpy sequences, substantial camera angle variations and large-distances from dynamic objects can lead to the omission of objects in point clouds, primarily due to the limited range of the depth camera. As a result, this deficiency in feature matching slightly impacted the performance of our approach.  Fig. \ref{fig:Plot_Results} represents the 2D projections of 3D trajectories of 3DS-SLAM and ORB-SLAM2. In both high and low dynamic sequences, our proposed 3DS-SLAM trajectories closely match the ground truth, whereas the trajectory estimated by ORB-SLAM2 exhibits a significant deviation from the ground truth. Overall, 3DS-SLAM demonstrates a substantial average improvement of $98.01\%$ over ORB-SLAM2 in dynamic sequences from the TUM RGB-D dataset.

\begin{figure}[t]
    \centering
    \centerline{\includegraphics[width=\linewidth]{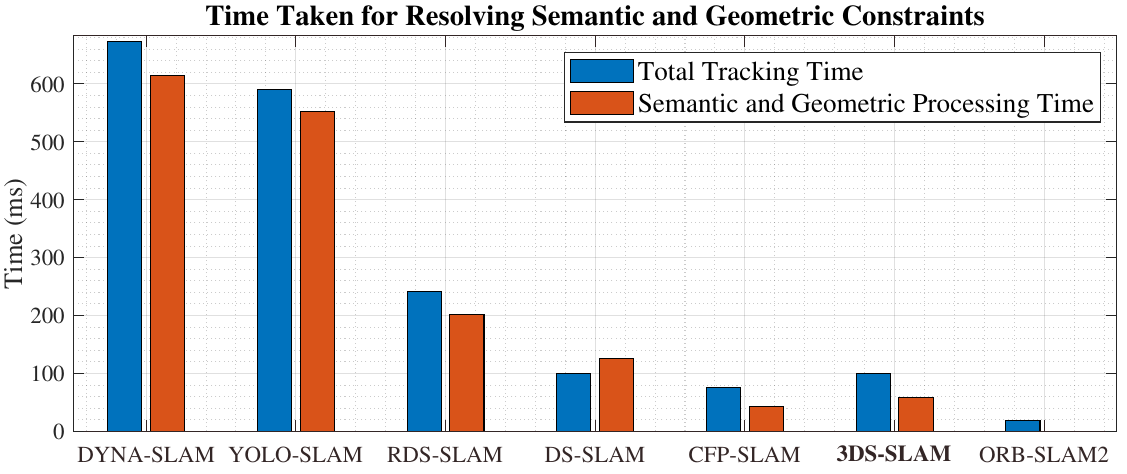}}
    \captionsetup{font=small}
    \caption {The comparison of execution time with existing SLAMs.}
    \label{fig:realtime}
    \vspace{-8mm}
\end{figure}

Real-time performance and computational efficiency are crucial for responsive and accurate visual SLAM frameworks. Figure \ref{fig:realtime} illustrates a comparison of processing times among existing architectures, where processing for semantic and geometric constraints includes 3D object detection and dynamic feature filtering. Pose estimation and ORB feature extraction contribute to overall tracking duration. While DYNA-SLAM and YOLO-SLAM exhibit strong tracking capabilities, they suffer from extended processing times due to the use of Mask R-CNN and Darknet19-YOLOv3, respectively. DS-SLAM and CFP-SLAM process frames rapidly but struggle with sequences featuring rapid camera rotations. In contrast to existing SLAM systems, 3DS-SLAM not only meets real-time requirements but also maintains high accuracy levels. To enhance its computational efficiency, we've implemented key measures: 1) Parallel execution of semantic and geometric threads with ORB feature extraction for consecutive frames. 2) Point-cloud preprocessing to eliminate unnecessary data, leading to improved speed and accuracy.

\subsection{Discussion}
In real-time scenarios, existing 2D visual SLAM frameworks face challenges, notably regarding the detection of missing objects \cite{maharani2019enhancement} and localizing dynamic objects. In dynamic environments, factors like object motion, partial object visibility within the camera's field of view, image blurring, varying lighting conditions,  and unique camera angles due to rotation pose significant hurdles for object detection in critical robotic applications. Consequently, there is a need for the development of approaches for missing object detection and dynamic object localization, which are explicitly applied to object detectors. However, these approaches significantly increase the computational time required for existing visual SLAM systems. For instance, CFP-SLAM addresses the challenge of missing objects by explicitly incorporating extended Kalman filter and Hungarian algorithms with YOLOv5 but lacks computational efficiency. In contrast, 3DS-SLAM primarily aims to solve the missing dynamic object problem in visual SLAMs by implicitly combining part-aware object localization with object detection.

Incorporating point clouds into the framework improves dynamic object localization compared to manual parametric fusion of depth maps with RGB images as seen in existing visual SLAM practices \cite{wu2022yolo}, \cite{chen2016transforming}. It also overcomes limitations associated with 2D object detection due to environmental factors and critical robotic environments. The 3DS-SLAM system offers several advantages and future potential, including enhanced 3D scene understanding for object recognition, increased robustness in challenging lighting and texture conditions, and effective handling of under-hanging structures like tables and beds.

\section{CONCLUSION}
In this research work, we present a novel approach to visual SLAM employing a 3D hybrid transformer architecture tailored for highly dynamic environments. This investigation extensively contributes to enhancing the efficacy of visual SLAM systems through the incorporation of 3D scene comprehension. Rigorous assessments demonstrate that our algorithm attains superior localization accuracy across a wide spectrum of scenarios, spanning both low and high dynamic environments, while also exhibiting commendable real-time performance. In future, we will mainly aim to design a light-weight storage format for reconstructed point cloud map, which can be extensively used for precise robotic manipulation and navigation. 

\balance

{\small
\bibliography{references}
}

\end{document}